\documentclass[letterpaper, 10 pt, conference]{ieeeconf}

\IEEEoverridecommandlockouts
\usepackage{cite}
\usepackage{amsmath,amssymb,amsfonts}
\usepackage{algorithmic}
\usepackage{graphicx}
\usepackage{textcomp}
\usepackage{comment}
\def\BibTeX{{\rm B\kern-.05em{\sc i\kern-.025em b}\kern-.08em
    T\kern-.1667em\lower.7ex\hbox{E}\kern-.125emX}}
\usepackage{subfig}
\usepackage{tabularx}
\usepackage{wasysym}
\usepackage{booktabs}
\usepackage{graphicx} 
\usepackage{listings} 
\usepackage{multirow}
\usepackage[table]{xcolor}
\usepackage{array}
\usepackage{subcaption}
\usepackage{graphicx}
\usepackage{pdfpages}
\usepackage{capt-of} 
\usepackage{amsmath} 
\usepackage{hyperref} 
\usepackage{balance} 
\usepackage{siunitx} 

\usepackage{longtable}
\usepackage{amsmath, amssymb}

\title{\bf Natural Selection via Foundation Models  for Soft Robot Evolution}
\author{
Changhe Chen* \quad Xiaohao Xu${}^{\dagger}$* \quad Xiangdong Wang* \quad Jiaqi Wang \quad  Xiaonan Huang
\thanks{$\dagger$ Project Lead \quad $*$ Equal Contribution}
\thanks{The authors are with  University of Michigan-Ann Arbor, Ann Arbor, MI 48109 USA (E-mail: \{changhec, xiaohaox,  sheldonw, wangjq, xiaonanh\}@umich.edu).}
\thanks{${}^{1}$Code available at \href{https://github.com/robocrafterqa/robocrafterqa_code}{https://github.com/robocrafterqa/robocrafterqa\_code}}
}

\begin{document}

\maketitle

\begin{abstract}
Designing soft robots is a complex and iterative process that demands cross-disciplinary expertise in materials science, mechanics, and control, often relying on intuition and extensive experimentation. While foundation models, especially Large Language Models (LLMs), have demonstrated impressive reasoning abilities, their capacity to conduct embodied design remains largely unexplored. This paper introduces \textbf{\texttt{RoboCrafter-QA}}, a novel benchmark to evaluate whether LLMs can learn representations of soft robot designs that effectively bridge the gap between high-level task descriptions and low-level morphological and material choices. \texttt{RoboCrafter-QA} leverages the EvoGym simulator to generate a diverse set of soft robot design challenges, spanning robotic locomotion, manipulation, and balancing tasks. Our experiments with SOTA multi-modal LLMs reveal that while these models exhibit promising capabilities in learning design representations, they struggle with fine-grained distinctions between designs with subtle performance differences. To overcome these limitations, we finetune an efficient, open-source LLM that achieves SOTA performance on our benchmark, demonstrating superior capabilities in both design selection and direct generation of high-performing robot morphologies. Furthermore, we construct a physical replica of the modular soft robot and demonstrate a strong sim-to-real correlation, validating that superior benchmark performance has the potential to translate to effective real-world design selection. Our full system will be open-sourced to foster this exciting direction.
\end{abstract}

\section{Introduction}

Soft robots offer distinct advantages over traditional rigid-bodied systems, particularly in complex, unstructured, and dynamic environments, where their inherent compliance enables safer and more adaptable interactions \cite{Lipson2013}.  The compliance, adaptability, and distributed actuation of soft robots unlock new possibilities in applications ranging from minimally invasive surgery to flexible manufacturing and exploration \cite{Cianchetti2018, Huang2022, WALKER2019335}.  However, these same characteristics introduce significant design challenges for soft robots. Unlike rigid robots with well-defined kinematic chains, soft robots possess virtually infinite degrees of freedom, exhibit non-linear material properties, and necessitate intricate coordination of morphology, actuation, and control.  This complexity makes soft robot design a highly challenging multidisciplinary problem, traditionally relying on expert intuition, iterative prototyping, and computationally expensive simulations.

\begin{figure}[t] 
    \centering\includegraphics[width=0.48\textwidth]{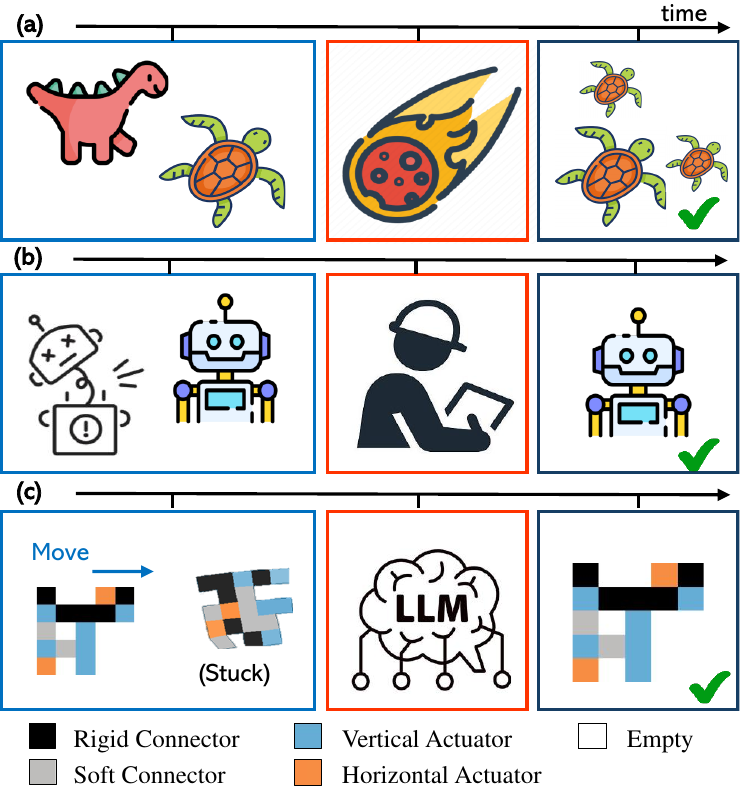} 
    \caption{\textbf{Conceptual comparison of  natural creature and artificial robot \textit{evolution} paradigms}. (\textbf{a}) Evolution-driven emergence of creatures in nature through selective pressures; (\textbf{b}) Traditional human-engineered robot design guided by intuition and expertise; (\textbf{c}) AI-driven robot design selection, where foundation models 
    (\textit{e.g.}, {Large Language Models) act as a `\textit{natural selector}' } for robot design evolution (\textit{e.g.}, modular robots shown in the figure). This shift highlights the transition from biological evolution to human-driven engineering and finally to AI-empowered  selection.}
    \label{fig:teaser}\vspace{-1mm}
\end{figure}

Fig.~\ref{fig:teaser} illustrates a conceptual shift in the paradigms of creature and robot design.  While biological evolution (Fig.~\ref{fig:teaser}a) and human-engineered design (Fig.~\ref{fig:teaser}b) have driven progress, they are inherently slow and limited by human cognitive capacity.  The emergence of foundation models, particularly Large Language Models (LLMs), presents a new opportunity: LLMs as a form of `natural selector' for robot design (Fig.~\ref{fig:teaser}c).  LLMs, trained on vast datasets that encompass text and code, have demonstrated impressive capabilities in the understanding, generation, and even reasoning of natural languages. As discussed in \cite{stella2023llms}, LLMs have the potential to revolutionize the robotic design process by serving as intelligent co-designers, however, a comprehensive and quantitative analyses and understanding of LLM for robot design is still underexplored. This raises our fundamental research question: \textbf{Can LLMs leverage their latent understanding for the design of embodied soft robots, bridging the gap between high-level task descriptions and low-level morphological choices?}

\begin{figure*}[t] 
    \centering
    \includegraphics[width=\textwidth]{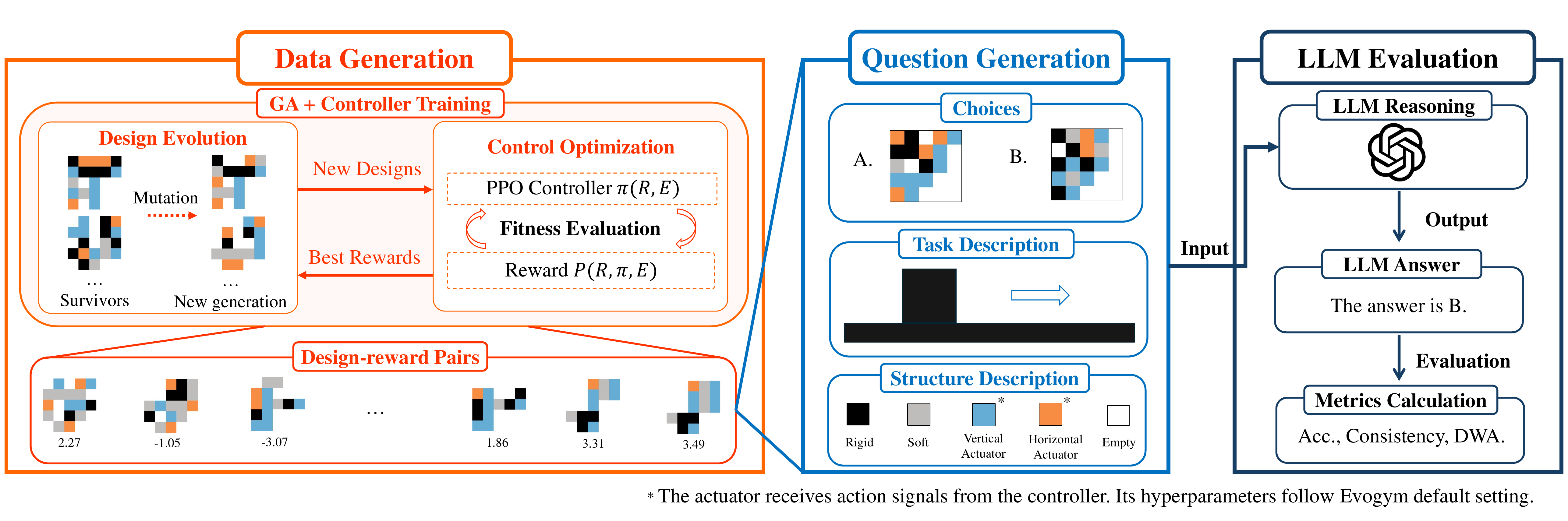} 
    \caption{{\textbf{\texttt{RobotCrafter-QA}}} robot design data generation and LLM evaluation pipeline. }
    \label{fig:pipeline}\vspace{-0mm}
\end{figure*}

Existing LLM benchmarks are ill-suited for this problem. They focus on general-purpose language, code, or non-embodied visual reasoning~\cite{Chen2021ScienceQA,Nguyen2022ScienceQA,Singh2023ScienceQA,BIGBench2022}, failing to assess an LLM's ability to reason about physical systems. Critically, they do not test the \textbf{embodied design intelligence} required for soft robot creation—namely, the complex interplay between a robot's morphology, its environment, and the task it must perform.
To address this gap, we introduce \textbf{\texttt{RoboCrafter-QA}}, a benchmark for evaluating the embodied design intelligence of multimodal LLMs (pipeline in Fig. \ref{fig:pipeline}). The benchmark is built using the EvoGym simulator~\cite{bhatia2021evolution} to generate a large-scale dataset of (environment, task, soft robot design, performance) tuples. The core task is a Question-Answer (QA) challenge: given a textual description of a task and environment, an LLM must select the better-performing robot design from a pair. This format directly tests the LLM's ability to reason about the complex interplay between morphology, materials, and performance, moving beyond simple pattern matching.

Our benchmarking results, using state-of-the-art LLMs, reveal both the potential and the limitations of current models in this challenging domain. We find that while LLMs can learn to distinguish between significantly different designs, they struggle with fine-grained distinctions, highlighting the need for further advancements in representation learning for embodied design. To push beyond these limitations, we demonstrate that an efficient, open-source LLM can be finetuned on \textbf{\texttt{RoboCrafter-QA}} to achieve state-of-the-art design selection. We show this model not only surpasses SOTA models in the selection task but can also be prompted to \textit{generate} novel, high-performing robot morphologies directly. Finally, to close the loop and validate the practical utility of our benchmark, we construct a physical modular robot system. Our real-world experiments confirm a strong sim-to-real correlation, demonstrating that the design principles learned by our model in simulation translate to effective performance in a physical robot.

This paper makes the following key contributions:
\begin{enumerate}
    \item We introduce \textbf{\texttt{RoboCrafter-QA}}, a novel multi-modal question-answering benchmark that assesses an LLM's ability to reason about the interplay between task requirements and soft robot design.
    \item We provide a comprehensive evaluation of state-of-the-art M-LLMs on \textbf{\texttt{RoboCrafter-QA}}, revealing key insights into their current capabilities and limitations in embodied design reasoning.
    \item We finetune an efficient, open-source M-LLM that establishes a new state-of-the-art on our benchmark, demonstrating significantly stronger robot design selection and reasoning capabilities.
    \item We construct a physical replica of the modular soft robot and demonstrate a strong sim-to-real correlation, validating that superior benchmark performance translates to effective real-world design selection.
\end{enumerate}

\begin{figure*}[t] 
    \centering
    \includegraphics[width=\textwidth]{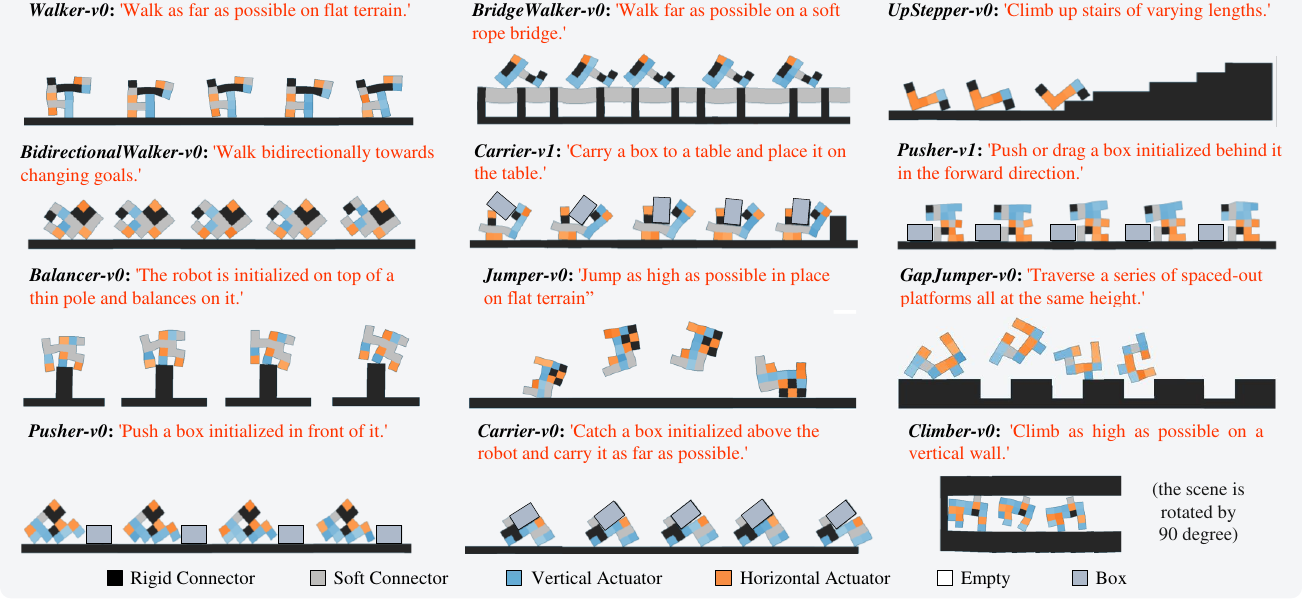} 
    \caption{Overview of robot design tasks curated from EvoGym~\cite{bhatia2021evolution} for the {\textbf{\texttt{RoboCrafter-QA}}} benchmark.} \vspace{-0mm}
    \label{fig:all_tasks}
\end{figure*}

\begin{figure*}[t] 
     \centering
    \includegraphics[width=\textwidth]{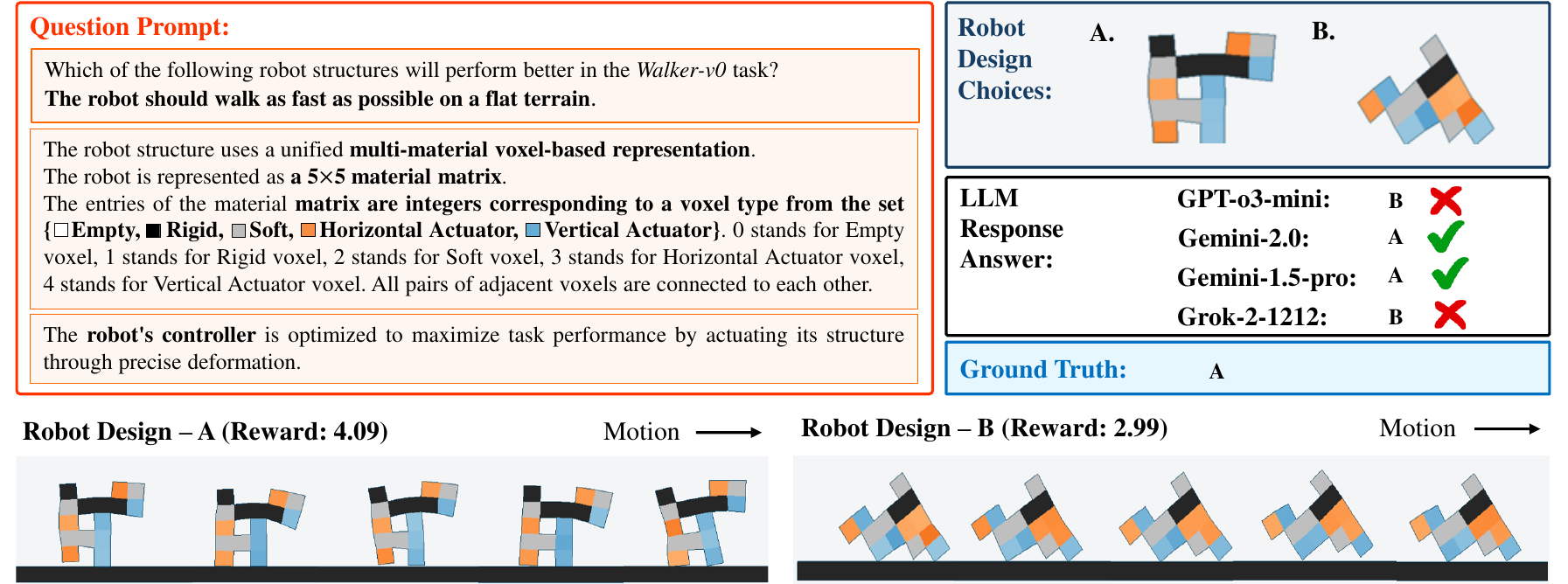} 
    \caption{{Example \textbf{\texttt{RoboCrafter-QA}} question and LLM responses.} This figure illustrates a sample question from the benchmark, showing the two robot designs (color-coded from input matrices), the question prompt, and the LLM responses with the ground truth. The simulated locomotion of the robot designs are illustrated at the bottom.}
    \label{fig:qa_diffdesign} \vspace{-1mm}
\end{figure*}

\section{Related Work}

\subsection{Soft Modular Robot Design and Optimization}
Soft modular robots leverage compliance and distributed actuation (e.g., pneumatics \cite{robertson2021soft}, SMAs \cite{huang2019highly}, EAPs \cite{gu2021soft}) to achieve adaptability and dynamic morphology, unlike their rigid counterparts \cite{Smith2021SoftModular,Li2022ModularSoft,Chen2023Reconfigurable,Doe2021SoftRobots}. Research has focused on reconfigurability, including lattice-based adaptation to external forces \cite{Zhang2023MorphAdapt}, flexible attachment mechanisms \cite{Li2023SpatialDesign}, and real-time, sensor-driven reconfiguration \cite{Nguyen2023ReconfigSoft,Patel2024DynamicMorphology}.

Optimizing these robots is challenging due to their nonlinear dynamics. Early approaches pioneered evolutionary algorithms for automatic design \cite{10.1145/2661735.2661737}, a method later used to create \textit{xenobots} from biological cells \cite{Kriegman2020Evolving}. More recent work leverages generative AI, such as diffusion models with differentiable physics \cite{Wang2023DiffuseBot} and hybrid data-driven/physics-based methods \cite{Garcia2025OptimizedSoft}, for automated morphology discovery. Our work contributes by benchmarking LLMs for automated design selection, using the EvoGym simulator to systematically probe their reasoning on morphology optimization under realistic physical constraints.

\subsection{LLM Evaluation Benchmarks}
General benchmarks like MMLU \cite{MMLU2023} and C-Eval \cite{Li2023CEval} test broad linguistic and reasoning skills. In response to rapid model evolution, specialized benchmarks such as ScienceQA \cite{Nguyen2022ScienceQA,Singh2023ScienceQA} and MMLU-pro \cite{Rizzo2023MMLUPro,Li2023MMLUPro} have emerged to test deep, domain-specific knowledge, though they remain focused on textual understanding.

These text-centric benchmarks overlook embodied intelligence, where physical constraints and morphology are critical. LLMs are known to struggle with robotics-related physical reasoning \cite{Li2023LLMZeroShotExplorers}, highlighting the need for specialized assessments that bridge high-level planning with real-world physical constraints \cite{Huang2023RoboticLangEval,Yao2025PhysicalReasoning}. Existing benchmarks fail to measure how LLMs integrate task goals, environmental conditions, and morphology. We introduce \textbf{{\texttt{RoboCrafter-QA}}}, a structured benchmark to assess how  LLMs reason about soft robot design choices. \textbf{\texttt{RoboCrafter-QA}} moves beyond purely textual evaluation to provide a comprehensive assessment of embodied intelligence.

\section{RobotCrafter-QA Benchmark}

\subsection{Benchmark Overview}
\begin{figure}[t] 
    \centering
\includegraphics[width=0.48\textwidth]{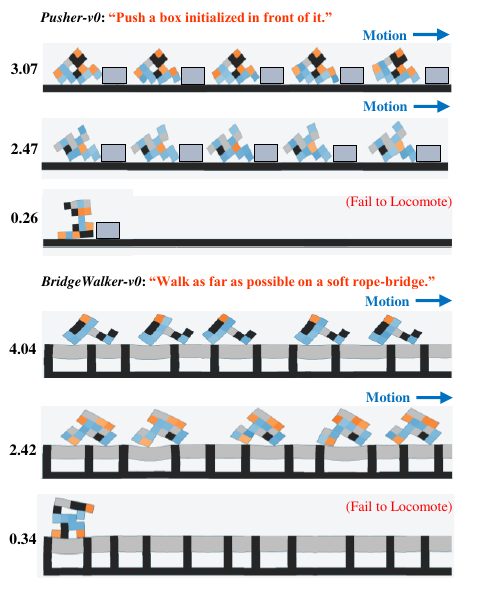} 
 \vspace{-5mm}
    \caption{Example of robot designs with varied rewards.}
    \label{fig:reward_design}  \vspace{-2mm}
\end{figure}

The \textbf{\texttt{RoboCrafter-QA}} benchmark is designed to rigorously evaluate the embodied design reasoning capabilities of LLMs in the context of soft robot design over 12 tasks as illustrated in Fig. \ref{fig:all_tasks}. It comprises a curated dataset of Question-Answer (QA) pairs, $\{(Q_i, G_i)\}_{i=1}^{N}$, generated using the EvoGym simulator. Each question $Q_i = (D_{Q_i}, R_{1_i}, R_{2_i})$ presents a LLM with a textual prompt $D_{Q_i}$ and two candidate robot designs, $R_{1_i}$ and $R_{2_i}$, represented as voxel-based material matrices. The ground truth answer, $G_i \in \{A, B\}$, indicates the robot design that achieves the higher reward, $P(R, \pi, E)$, in the simulated environment $E$. The benchmark encompasses a diverse range of task environments, $E = (D, T, C)$, including locomotion, object manipulation, and climbing, each characterized by a textual description $D$, a task objective $T$, and environmental constraints $C$. An example of a question $Q_i$ and its components is illustrated in Fig. \ref{fig:qa_diffdesign}. The difficulty of each question is modulated by the difference in performance rewards, $|P(R_{1_i}, \pi, E) - P(R_{2_i}, \pi, E)|$, between the candidate robot designs, enabling a nuanced assessment of LLMs' ability to discern subtle design trade-offs. An example of various robot designs and their corresponding rewards for a pushing task and a bridge walking task is shown in Fig.~\ref{fig:reward_design}. The dataset is partitioned into \textit{Easy} and \textit{Hard} subsets based on this performance difference, providing a comprehensive evaluation of LLMs' embodied design reasoning across varied complexity.

\renewcommand{\arraystretch}{1.0} 

\subsection{Data Generation}

\noindent\textbf{Design Space.} Robot morphologies are evolved within a $5 \times 5$ voxel-based design space, where each voxel denotes a material type: empty (0), rigid (1), soft (2), horizontal actuator (3), or vertical actuator (4). Structural integrity is enforced by maintaining fully connected components. Thus, a robot design $R$ is represented by a material matrix $M \in \{0, 1, 2, 3, 4\}^{5 \times 5}$.

\noindent\textbf{Evolution.} The evolutionary process begins with an initial population of 30 unique robot designs, generated randomly. Each robot $R$ is evaluated in a designated Evogym environment $E$ using a Proximal Policy Optimization (PPO)-trained~\cite{schulman2017proximal} controller $\pi(R, E)$, which controls the action signals applied on the actuator voxels. The controller is trained in the Evogym simulation to optimize the performance of the given robot structure. The fitness metric is the best reward, $P(R, \pi, E)$, reflecting per-task performance. We refer the reader to \cite{bhatia2021evolution} for a detailed description of the parameter settings, including voxel specifications and environmental conditions.

\noindent\textbf{Selection and Mutation.} The top 50\% of robots per generation are retained, while the remaining are replaced via mutation-based offspring generation. Mutations modify voxel properties at a low probability (about 10\%), ensuring design diversity. Novelty constraints are enforced before acceptance to maintain uniqueness.

\noindent\textbf{Termination and Data Storage.} The evolutionary process continues with a constant population size of 30 until 3000 robots have been evaluated or an alternative termination criterion is met. The final population, along with its morphological data $M$ and performance data $P$, is archived for further analysis.

\begin{table}[t]
    \renewcommand{\arraystretch}{1.2}
    \centering
    \setlength\tabcolsep{1mm}
    \caption{Statistics of question sets across different environments, including the number of questions, reward difference range, maximum reward, and minimum reward.}
    \resizebox{\columnwidth}{!}{%
    \begin{tabular}{l|c|c|cc}
        \toprule
        \textbf{Task Name} & \textbf{\# Questions} & \textbf{Reward Diff} & \textbf{Max Reward} & \textbf{Min Reward} \\
        \midrule
        \textit{Walker-v0} & \multirow{12}{*}{3,000} & 0.74 to 2.89 & 4.09 & -1.21 \\
        \textit{BridgeWalker-v0} &  & 0.39 to 3.69 & 4.04 & -1.49 \\
        \textit{BidirectionalWalker-v0} &  & 0.31 to 4.16 & 1.81 & -3.69 \\
        \textit{UpStepper-v0} &  & 0.08 to 2.51 & 1.64 & -1.27 \\
        \textit{GapJumper-v0} &  & 0.08 to 2.62 & 2.28 & -0.77 \\
        \textit{Carrier-v0} &  & 0.53 to 3.50 & 2.81 & -2.86 \\
        \textit{Carrier-v1} &  & 0.90 to 6.62 & 3.60 & -5.85 \\
        \textit{Pusher-v0} &  & 0.49 to 3.07 & 3.07 & -1.66 \\
        \textit{Pusher-v1} &  & 0.04 to 1.39 & 0.21 & -1.33 \\
        \textit{Climber-v0} &  & 0.01 to 0.19 & 0.16 & -0.18 \\
        \textit{Jumper-v0} &  & 0.13 to 8.07 & 0.04 & -8.44 \\
        \textit{Balancer-v0} &  & 0.12 to 1.05 & 0.13 & -1.41 \\
        \midrule
        \textbf{Summary} & 36,000 & 0.01 to 8.07 & 4.09 & -8.44 \\
        \bottomrule
    \end{tabular}
    }
    \label{tab:env_stats}
    \vspace{-1mm}
\end{table}

\subsection{Question Generation}

\noindent\textbf{Choice Construction.} An automated pipeline transforms experimental data into structured multi-choice questions for LLM evaluation. Robot records are sorted by reward $P$ and partitioned into pairs, forming questions that require models to identify the higher-performing structure.

\noindent\textbf{Context Integration.} Each question includes environment-specific objectives, such as \textit{`Climb as high as possible on a vertical wall’} or `\textit{Walk as far as possible on flat terrain}' to provide necessary context.

\noindent\textbf{Structured Representation.} Each question $Q_i$ consists of:
\begin{itemize}
    \item A $5 \times 5$ voxel grid representation $M$ detailing material types (rigid, soft, actuator, or empty).
    \item Information on the actuation mechanism $\pi$ influencing movement.
    \item A letter $G_i \in \{A, B\}$ indicating the superior robot based on empirical performance $P$.
\end{itemize}


\begin{table*}[t]
\centering
\caption{Accuracy and Consistency metrics across different LLMs for various robotic design tasks.}
\label{tab:llm_environment_results}
\setlength\tabcolsep{2mm}
\resizebox{\textwidth}{!}{
\begin{tabular}{ll|cc|cccc|cccc|cccc|cccc}
\toprule
\multirow{4}{*}{\shortstack{\textbf{Category}}} & \multirow{4}{*}{\shortstack{\textbf{Setting}}}
& \multicolumn{2}{c|}{\textbf{Heuristic}}
& \multicolumn{4}{c|}{\textbf{Qwen-0.5-Instruct}}
& \multicolumn{4}{c|}{\textbf{GPT5}}
& \multicolumn{4}{c|}{\textbf{Gemini-2.5-Pro}}
& \multicolumn{4}{c}{\textbf{Grok-4}} \\
\cmidrule(lr){3-4} \cmidrule(lr){5-8} \cmidrule(lr){9-12} \cmidrule(lr){13-16} \cmidrule(lr){17-20}
& & \multicolumn{2}{c|}{\textbf{Accuracy}}
& \multicolumn{2}{c}{\textbf{Accuracy}} & \multicolumn{2}{c|}{\textbf{Consistency}}
& \multicolumn{2}{c}{\textbf{Accuracy}} & \multicolumn{2}{c|}{\textbf{Consistency}}
& \multicolumn{2}{c}{\textbf{Accuracy}} & \multicolumn{2}{c|}{\textbf{Consistency}}
& \multicolumn{2}{c}{\textbf{Accuracy}} & \multicolumn{2}{c}{\textbf{Consistency}} \\
\cmidrule(lr){3-4} \cmidrule(lr){5-6} \cmidrule(lr){7-8} \cmidrule(lr){9-10} \cmidrule(lr){11-12} \cmidrule(lr){13-14} \cmidrule(lr){15-16} \cmidrule(lr){17-18} \cmidrule(lr){19-20}
& & \textbf{Easy} & \textbf{Hard}
& \textbf{Easy} & \textbf{Hard} & \textbf{Easy} & \textbf{Hard}
& \textbf{Easy} & \textbf{Hard} & \textbf{Easy} & \textbf{Hard}
& \textbf{Easy} & \textbf{Hard} & \textbf{Easy} & \textbf{Hard}
& \textbf{Easy} & \textbf{Hard} & \textbf{Easy} & \textbf{Hard} \\
\midrule

\multirow{6}{*}{\textbf{Locomotion}}
& \textit{Walker-v0} & 49.61 & 54.03 & 39.17 & 37.83 & 39.00 & 38.33 & 51.34 & 49.34 & \textbf{76.67} & \textbf{78.67} & 48.33 & 49.84 & 66.67 & 69.67 & \textbf{60.33} & \textbf{56.84} & 72.00 & 70.33 \\
& \textit{BridgeWalker-v0} & 45.53 & 46.59 & 37.67 & 38.67 & 40.00 & 42.67 & 54.33 & 52.84 & 70.67 & \textbf{71.67} & 61.16 & 52.33 & 71.67 & 66.00 & \textbf{69.33} & \textbf{56.66} & \textbf{74.67} & 68.00 \\
& \textit{BidirectionalWalker-v0} & 48.79 & 51.91 & 36.33 & 36.50 & 37.33 & 38.00 & 50.66 & \textbf{53.84} & 73.33 & \textbf{77.67} & \textbf{51.66} & 47.50 & 70.00 & 69.67 & 48.50 & 51.50 & \textbf{76.33} & 77.00 \\
& \textit{UpStepper-v0} & 57.14 & 53.06 & 41.83 & 41.17 & 48.33 & 41.00 & \textbf{67.17} & 51.66 & \textbf{82.33} & \textbf{85.33} & 61.66 & 51.66 & 69.33 & 68.67 & 64.83 & \textbf{56.50} & 78.33 & 76.33 \\
& \textit{GapJumper-v0} & 51.29 & 50.66 & 37.67 & 38.17 & 35.67 & 36.33 & 58.00 & 51.84 & \textbf{72.67} & \textbf{74.33} & 60.17 & 57.17 & 65.67 & 67.00 & \textbf{64.67} & \textbf{57.66} & 68.67 & 66.67 \\
\cmidrule{2-20}
& \textbf{{Average}} & 50.47 & 51.25 & 38.53 & 38.47 & 40.07 & 39.27 & 56.30 & 51.90 & \textbf{75.13} & \textbf{77.53} & 56.60 & 51.70 & 68.67 & 68.20 & \textbf{61.53} & \textbf{55.83} & 74.00 & 71.67 \\
\midrule

\multirow{5}{*}{\shortstack{\textbf{Object} \\ \textbf{Manipulation}}}
& \textit{Carrier-v0} & 45.07 & 51.02 & 38.67 & 37.17 & 33.33 & 31.33 & \textbf{69.16} & 57.84 & \textbf{82.33} & \textbf{79.67} & 60.34 & 51.83 & 68.00 & 70.33 & 68.34 & \textbf{58.34} & 77.33 & 72.00 \\
& \textit{Carrier-v1} & 40.22 & 43.66 & 34.50 & 35.00 & 40.67 & 36.00 & 65.34 & 59.83 & \textbf{81.33} & \textbf{75.67} & \textbf{67.16} & \textbf{61.33} & 77.00 & 74.67 & 65.00 & 57.84 & 72.67 & 71.00 \\
& \textit{Pusher-v0} & 45.26 & 57.19 & 38.33 & 41.00 & 42.67 & 42.33 & 30.34 & 45.84 & \textbf{74.00} & \textbf{76.33} & 40.50 & 50.66 & 73.67 & 76.00 & \textbf{58.50} & \textbf{59.66} & 69.00 & 72.67 \\
& \textit{Pusher-v1} & 56.01 & 52.30 & 39.33 & 38.50 & 40.67 & 39.33 & 49.16 & 48.84 & 72.33 & 79.00 & 51.00 & 54.67 & 68.00 & 74.00 & \textbf{53.84} & \textbf{56.50} & \textbf{76.33} & \textbf{81.00} \\
\cmidrule{2-20}
& \textbf{{Average}} & 46.64 & 51.04 & 37.71 & 37.92 & 39.33 & 37.25 & 53.50 & 53.09 & \textbf{77.50} & \textbf{77.67} & 54.75 & 54.62 & 71.67 & 73.75 & \textbf{61.42} & \textbf{58.09} & 73.83 & 74.17 \\
\midrule

\multirow{4}{*}{\shortstack{\textbf{Climbing} \& \\ \textbf{Balancing}}}
& \textit{Climber-v0} & 54.04 & 55.82 & 37.00 & 37.00 & 37.67 & 36.33 & 51.50 & 49.84 & 79.00 & 75.67 & 55.50 & 57.00 & 71.67 & 75.33 & \textbf{57.34} & \textbf{57.67} & \textbf{79.33} & \textbf{77.33} \\
& \textit{Jumper-v0} & 18.13 & 33.42 & 36.00 & 37.00 & 42.67 & 41.00 & 9.84 & 31.50 & \textbf{91.67} & \textbf{82.33} & 16.84 & 39.67 & 86.33 & 78.67 & \textbf{42.16} & \textbf{44.34} & 72.33 & 68.00 \\
& \textit{Balancer-v0} & 53.71 & 58.38 & 34.83 & 36.50 & 33.00 & 37.67 & 54.66 & 52.33 & 70.00 & \textbf{72.67} & 51.17 & \textbf{59.16} & 68.33 & 69.67 & \textbf{63.84} & 57.34 & \textbf{76.33} & 68.00 \\
\cmidrule{2-20}
& \textbf{{Average}} & 41.96 & 49.21 & 35.94 & 36.83 & 37.78 & 38.33 & 38.67 & 44.56 & \textbf{80.22} & \textbf{76.89} & 41.17 & 51.94 & 75.44 & 74.56 & \textbf{54.45} & \textbf{53.12} & 76.00 & 71.11 \\
\midrule

  \textbf{Average}
& & 47.07 & 50.67 & 37.61 & 37.88 & 39.25 & 38.36 & 50.96 & 50.46 & \textbf{77.19} & \textbf{77.42} & 52.12 & 52.73 & 71.36 & 71.64 & \textbf{59.72} & \textbf{55.90} & {74.44} & {72.36} \\
\bottomrule
\end{tabular}
}
\vspace{-3mm}
\end{table*}

\subsection{Benchmark Statistics}

Table \ref{tab:env_stats} provides a detailed overview of the \textbf{\texttt{RoboCrafter-QA}} benchmark dataset, showcasing its breadth and controlled difficulty across twelve distinct EvoGym tasks. Comprising a substantial 36,000 questions in total, with 3,000 questions dedicated to each environment, the benchmark ensures a robust evaluation. The `Reward Diff' column highlights a critical aspect of question difficulty, presenting the range of performance score differences between the paired robot designs. This range varies across environments, from subtle distinctions (e.g., Climber-v0 with 0.01 to 0.19) to more pronounced performance gaps (e.g., Jumper-v0 with 0.13 to 8.07), effectively modulating the challenge for LLMs.  Furthermore, the `Max Reward' and `Min Reward' columns contextualize the performance landscape within these environments, demonstrating the variability in achievable and poor robot designs that contribute to the question generation process. This statistical breakdown underscores the dataset's comprehensive nature, designed to rigorously challenge and differentiate the embodied design reasoning capabilities of multi-modal LLMs across a spectrum of task complexities and performance disparities.

\section{Experiments}
\subsection{Evaluation Metrics of Robot Design Selection}\label{eval_metrics}

\noindent\textbf{Accuracy.} It measures the proportion of model-generated responses that match the expected answers. Given $N$ test queries, accuracy $Acc$ is computed as:
\begin{equation}
Acc = \frac{1}{N} \sum_{i=1}^{N} \mathbb{I} ( O_i = G_i )
\end{equation}
where $O_i$ is the prediction, $G_i$ is the ground truth, and $\mathbb{I} (\cdot)$ is an indicator function returning 1 if responses match, 0 otherwise.  

\noindent\textbf{Consistency.} It evaluates the stability of a model by measuring the reproducibility of the response in multiple trials. Each query is presented three times, and the consistency  is:
\begin{equation}
Cons = \frac{1}{N} \sum_{i=1}^{N} \frac{\mathbb{I} ( O_{i}^{(1)} = O_{i}^{(2)} ) + \mathbb{I} ( O_{i}^{(1)} = O_{i}^{(3)} )}{2}
\end{equation}
where $O_{i}^{(1)}, O_{i}^{(2)}, O_{i}^{(3)}$ are outputs from three independent trials. Higher consistency indicates a more stable and deterministic model behavior.

\subsection{Benchmarking Results of Robot Design Selection}
We evaluate the most advanced available LLMs across 12 diverse environments, using metrics categorized into `\textit{Easy}' and `\textit{Hard}' tasks in Table \ref{tab:llm_environment_results}, together with a heuristic baseline method.

\vspace{0.5mm}

\noindent\textbf{Finding 1: Simple heuristics are insufficient.}
The actuator-count heuristic provides a non-trivial baseline (Avg. Acc: 47.07\% Easy, 50.67\% Hard), but its sub-51\% accuracy confirms simple metrics are insufficient for this complex evaluation. Grok-4 (Avg. Acc: 59.72\%/55.90\%) and Gemini-2.5-Pro (52.12\%/52.73\%) consistently outperform this baseline, capturing more complex structure-function relationships. Conversely, GPT5's accuracy (50.96\%/50.46\%) is on par with or worse than the heuristic, suggesting its high consistency stems from less effective internal biases.

\vspace{0.5mm}
\noindent\textbf{Finding 2: A clear hierarchy of model capabilities and trade-offs.}
A distinct performance hierarchy emerges. Grok-4 shows the strongest reasoning, leading in average accuracy (59.72\% Easy, 55.90\% Hard). Conversely, GPT5 achieves the highest consistency (77.19\%/77.42\%) but its accuracy ($\approx$51\%) is substantially lower, failing to reliably beat the heuristic. Gemini-2.5-Pro offers a balanced profile (Acc: $\approx$52\%, Cons: $\approx$71\%). Qwen-0.5-Instruct performs worst on both metrics, falling well below the baseline.

\vspace{0.5mm}
\noindent\textbf{Finding 3: Task-dependent performance reveals systematic model biases.}
Model performance is highly task-dependent, revealing specific model biases. The \textit{Jumper-v0} task is exceptionally challenging: GPT5 and Gemini-2.5-Pro achieve minimal accuracy (9.84\% and 16.84\%, respectively). Critically, this failure is paired with high consistency (91.67\% and 86.33\%), indicating a convergence on a suboptimal design, not random guessing. This suggests a misinterpretation of dynamic propulsion. In contrast, strong performance on \textit{Carrier-v0} and \textit{UpStepper-v0} (65-69\% accuracy) implies LLMs reason better about structured locomotion and balance than about impulsive dynamics.

\subsection{Ablation Study on Minor Prompt Design}

\begin{table}[t]
    \centering
    \caption{Ablation study results on prompt design.}
    \setlength\tabcolsep{2mm} 
    \resizebox{0.48\textwidth}{!}{
    \begin{tabular}{l|c|c|c|c}
        \toprule
        \textbf{Setting}  & \textbf{Qwen-0.5-Instruct} & \textbf{GPT5} & \textbf{Gemini-2.5-Pro} & \textbf{Grok-4} \\
        \midrule
        \textit{NoAct}  & \textbf{53.31}  & 47.62  & 50.79  & 54.72 \\
           
        \textit{NoEnv}  & 51.31  & 46.96  & 49.79  & 56.72 \\
        \textit{Worse}  & 26.61  & 42.66  & 47.45  & 58.38 \\
        \midrule
          
        \textbf{Full} &  37.61 & \textbf{50.96} & \textbf{52.12} & \textbf{59.72}  \\
        \bottomrule
    \end{tabular}
    }
    \label{tab:ablation_results}\vspace{-2mm}
\end{table}

We conduct an ablation study on prompt design to examine how different components of the textual input affect each model’s ability to reason about robot design quality. Since voxel-based actuator encoding is essential for mechanical interpretation, we focus the ablations on task description and actuator specification, and additionally include a logical inversion setting where the models are instructed to select the worse design instead of the better one, to assess robustness in decision reversal.

\noindent\textbf{Finding: Role of contextual cues and model capacity in prompt understanding.}
As shown in Table~\ref{tab:ablation_results}, removing key contextual information leads to varied effects across models. For GPT-5, Gemini-2.5-Pro, and Grok-4, the full prompt yields the highest accuracy ($\approx$51–60\%), confirming that complete task and environment cues are essential for consistent reasoning. In contrast, the smaller Qwen-0.5B shows unexpectedly higher performance when actuator or environment details are removed (\textit{NoAct}: 53.31\%), surpassing its own full-prompt accuracy (37.61\%). This inversion likely stems from its limited capacity and overfitting to surface correlations, allowing it to perform better when the prompt is simplified. Across all models, the incomplete-prompt setting consistently degrades performance, indicating that logical negation and reasoning reversal remain challenging even for larger LLMs, which tend to retain a strong bias toward selecting optimal designs. Overall, these results highlight that task and environment context are indispensable for high-capacity models, while smaller LLMs may benefit from prompt simplification to avoid confusion or over-conditioning.

\subsection{Finetuning LLM for Robot Design Selection}

\begin{figure}[t] 
    \centering
    \includegraphics[width=0.48\textwidth]{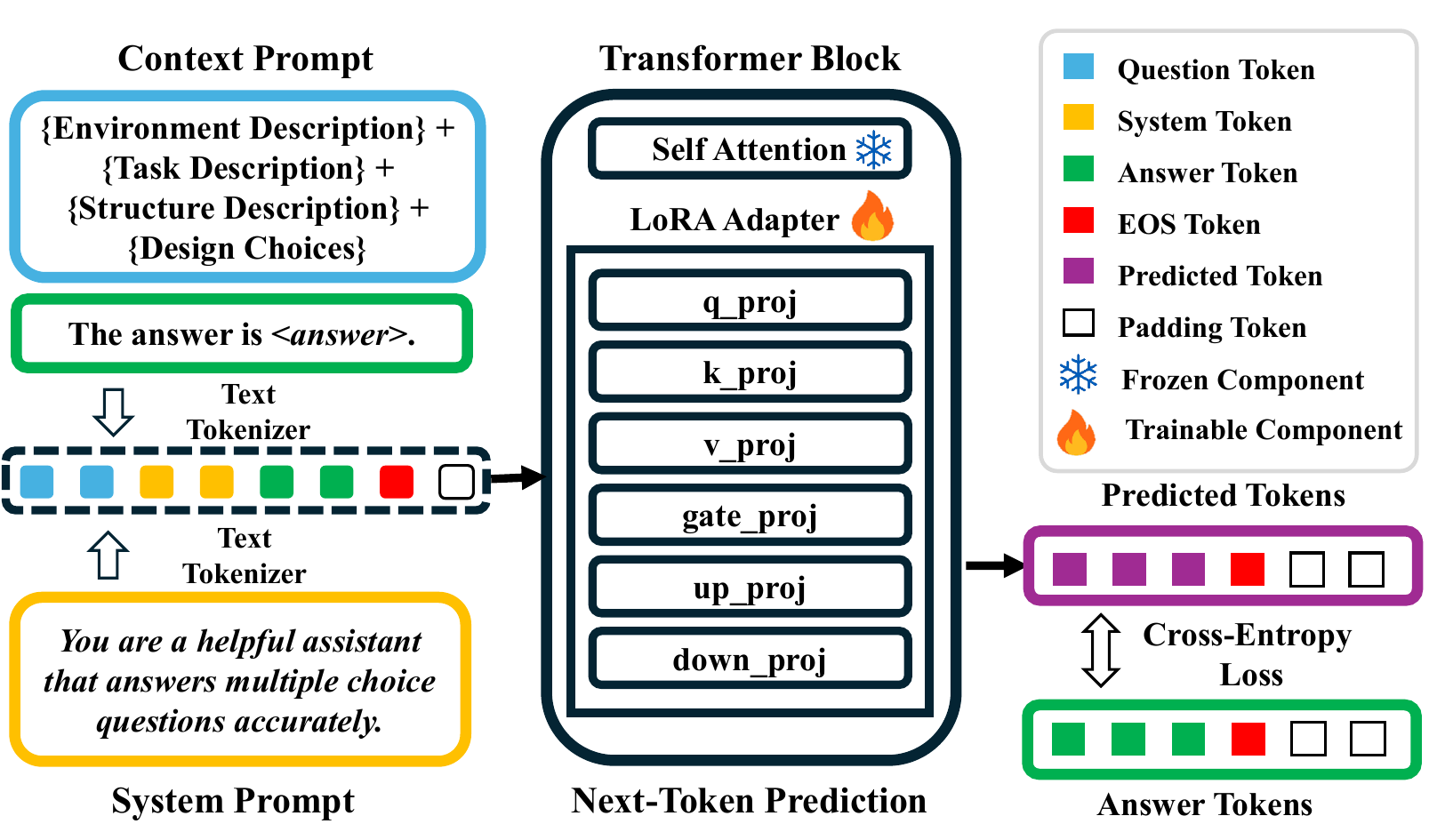} 
    \caption{LLM finetuning pipeline for soft robot design. }
    \label{fig:llm_pipeline}\vspace{-1mm}
\end{figure}

\begin{table}[t]
\centering
\caption{Finetuned Qwen-0.5B-Instruct performance across 12 environments for robot design selection.}
\label{tab:finetune_only_accuracy_consistency}
\setlength\tabcolsep{3mm} 
\resizebox{0.48\textwidth}{!}{ 
\begin{tabular}{l|cc|cc}
\toprule
\multirow{2}{*}{\textbf{Setting}} & 
\multicolumn{2}{c|}{\textbf{Accuracy}} & 
\multicolumn{2}{c}{\textbf{Consistency}} \\ 
\cmidrule{2-5}
& \textbf{Easy} & \textbf{Hard} & \textbf{Easy} & \textbf{Hard} \\ 
\midrule
\textit{Walker-v0}                & 99.50 & 95.00 & 99.67 & 95.00 \\
 \textit{BridgeWalker-v0}          & 100.00 & 96.00 & 100.00 & 97.00 \\
\textit{BidirectionalWalker-v0} & 77.83 & 69.67 & 58.67 & 55.67 \\
 \textit{UpStepper-v0}             & 97.00 & 85.50 & 95.33 & 79.00 \\
\textit{GapJumper-v0}             & 96.33 & 92.50 & 98.00 & 89.00 \\
 \textit{Carrier-v0}              & 99.67 & 92.00 & 99.33 & 89.33 \\
\textit{Carrier-v1}               & 96.17 & 91.33 & 95.33 & 90.67 \\
 \textit{Pusher-v0}                & 100.00 & 94.50 & 100.00 & 92.33 \\
\textit{Pusher-v1}                & 89.33 & 86.50 & 84.33 & 80.00 \\
 \textit{Climber-v0}               & 86.83 & 83.17 & 80.00 & 77.33 \\
\textit{Jumper-v0}                & 98.17 & 89.50 & 97.33 & 87.67 \\
 \textit{Balancer-v0}              & 87.33 & 84.00 & 79.67 & 81.00 \\
\midrule
 \textbf{Average}        & \textbf{94.01} & \textbf{88.31} & \textbf{90.64} & \textbf{84.50} \\
\bottomrule
\end{tabular}
}
\end{table}

\begin{table}[t]
    \centering
    \caption{Average and best rewards across 10 generated robot designs in 3 real-world evaluated environments.}
    \setlength\tabcolsep{0.8mm}
    \resizebox{0.48\textwidth}{!}{
    \begin{tabular}{l|cc|cc|cc|cc}
        \toprule
       \multirow{2}{*}{\shortstack{\textbf{Setting}}} & \multicolumn{2}{c|}{\textbf{GPT-5}} 
        & \multicolumn{2}{c|}{\textbf{Gemini2.5-Pro}} 
        & \multicolumn{2}{c|}{\textbf{Grok-4}} 
        & \multicolumn{2}{c}{\textbf{Ours}} \\
        \cmidrule{2-9}
        & Avg & Best & Avg & Best & Avg & Best & Avg & Best \\
        \midrule
        \textit{Walker-v0}  & 0.35 & 0.86 & 0.17 & 0.43 & 0.02 & 0.64 & \textbf{2.98} & \textbf{4.51} \\
          
        \textit{Pusher-v0}  & 0.15 & 0.57 & 0.20 & 0.73 & 0.01 & 0.10 & \textbf{0.69} & \textbf{2.64} \\
        \textit{Carrier-v0} & -0.36 & -0.22 & -0.26 & -0.04 & -0.53 & -0.37 & \textbf{0.79} & \textbf{1.90} \\
        \midrule
          
        \textbf{Average} & 0.05 & 0.40 & 0.04 & 0.37 & -0.17 & 0.12 & \textbf{1.48} & \textbf{3.02} \\
        \bottomrule
    \end{tabular}
    }
    \label{tab:llm_design_comparison}
\end{table}

To unleash the full potential of LLM model's understanding the structure of soft modular robot, we finetune Qwen2.5-0.5B-Instruct using a parameter-efficient Low-Rank Adaptation (LoRA) strategy~\cite{hu2022lora}. As illustrated in Fig.~\ref{fig:llm_pipeline}, all training data are organized by environment and difficulty, where each folder contains JSON files of question–answer (QA) pairs. Each sample is formatted in the Qwen chat style, consisting of a system prompt (`You are a helpful assistant that answers multiple choice questions accurately'), a user query describing the environment, task, and structure design, and an assistant response in the form `The answer is \textless\textit{answer}\textgreater.' The complete sequence, which combining system, user, and assistant messages, is tokenized and right-padded to a fixed length before being passed into the transformer.

During finetuning, LoRA adapters are injected into both the attention and feed-forward projections, specifically the layers \texttt{q\_proj}, \texttt{k\_proj}, \texttt{v\_proj}, \texttt{o\_proj}, \texttt{gate\_proj}, \texttt{up\_proj}, and \texttt{down\_proj}, while all other parameters remain frozen. The model is optimized under a causal language modeling objective that predicts the assistant’s next token given all preceding context. Gradients are propagated only through the LoRA components, allowing the model to efficiently learn instruction-level reasoning for accurate evaluation of soft-robot designs across diverse environments.

\begin{figure*}[t!] 
    \centering
    \includegraphics[width=\textwidth]{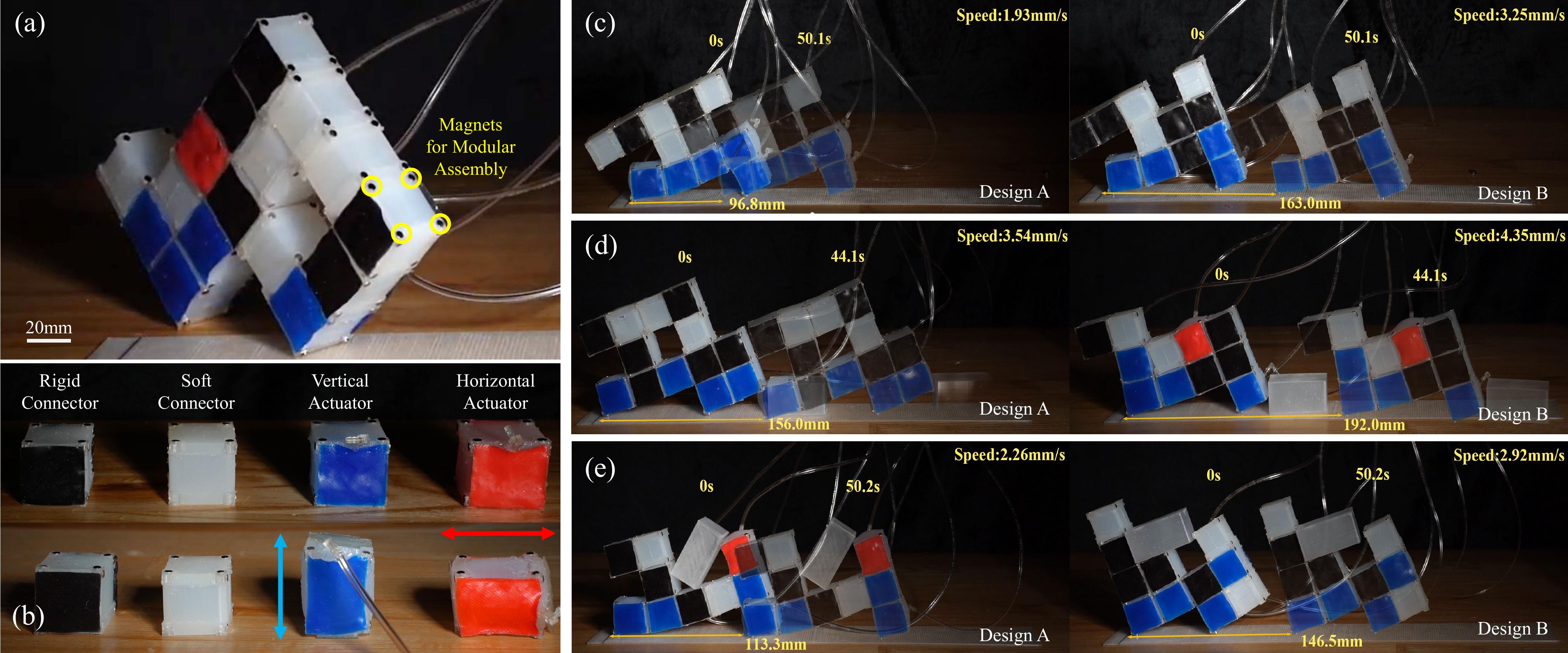} 
    \caption{\textbf{A real-world replica of soft modular robot designs in simulation.} (\textbf{a}) Real modular soft robot with (\textbf{b}) four basic modules assembled via tiny magnets integrated in each module.  Real-world speed comparison of modular soft robot designs for three tasks: (\textbf{c}) \textit{Walker-v0}, (\textbf{d}) \textit{Pusher-v0}, and (\textbf{e}) \textit{Carrier-v0}. Each task compares two modular designs from Table~\ref{tab:sim_vs_real_rewards}. }   \label{fig:real_modular_robot}
     \label{fig:real_modular_robot_demo}
\end{figure*}

The results of this finetuning process, detailed in Table~\ref{tab:finetune_only_accuracy_consistency}, demonstrate that this approach achieves exceptionally strong performance. The finetuned 0.5B model reaches a high average accuracy of 94.01\% on easy tasks and 88.31\% on hard tasks across all 12 environments. This high accuracy is complemented by strong average consistency of 90.64\% (Easy) and 84.50\% (Hard). Notably, the model achieves perfect 100\% accuracy and consistency on easy tasks for environments like BridgeWalker-v0 and Pusher-v0. This indicates that our parameter-efficient finetuning strategy is highly effective at enabling even a small model to understand and accurately evaluate complex robotic designs.

\begin{table}[t]
    \centering
    \caption{Comparison of simulation rewards and real-world metrics for  robot designs across three tasks. }\vspace{-1mm}
    \setlength\tabcolsep{1mm}
    \resizebox{0.48\textwidth}{!}{
    \begin{tabular}{l|l|c|c}
        \toprule
        \textbf{Setting} & \textbf{Design} & \textbf{Simulated Reward $\uparrow$} & \textbf{Real-World/\si{\milli\meter\per\second} $\uparrow$} \\
        \midrule
        \multirow{2}{*}{\textit{Walker-v0}} 
            & Design A & 3.17 & 1.93 \\
            & Design B & \textbf{3.26} & \textbf{3.25} \\
        \cmidrule(lr){1-4}
        \multirow{2}{*}{\textit{Pusher-v0}} 
            & Design A & 2.25 & 3.54 \\
            & Design B & \textbf{2.78} & \textbf{4.35} \\
        \cmidrule(lr){1-4}
        \multirow{2}{*}{\textit{Carrier-v0}} 
            & Design A & 0.95 & 2.26 \\
            & Design B & \textbf{1.32} & \textbf{2.92} \\
        \bottomrule
    \end{tabular}
    }
    \label{tab:sim_vs_real_rewards}\vspace{-2mm}
\end{table}

\subsection{LLM for Direct Robot Design}
After finetuning, we further evaluated the model’s generative capability by prompting it to directly design soft-robot structures instead of merely selecting between options. To ensure manufacturability and real-world verification in Sec.~\ref{sec:real-world}, we constrained the LLM to generate designs containing no more than five actuators across three selective environments.
As shown in Table \ref{tab:llm_design_comparison}, the finetuned Qwen-0.5B model (“Ours”) achieved the highest average and best rewards across all three environments: Walker-v0, Pusher-v0, and Carrier-v0, which largely outperforms existing general-purpose LLMs, including GPT-5, Gemini-2.5-Pro, and Grok-4. Our model obtained a mean average reward of 1.48, exceeding the next best baseline by over 10×, and a mean best reward of 3.02, demonstrating that the finetuned model not only understands design-quality distinction but can also generate effective and high-reward robot morphologies. 

These results confirm that instruction-level finetuning with structured QA data effectively transfers to real design generation, enabling compact LLMs to reason about and synthesize performant soft-robot structures across environments.

\subsection{Real-world Robot Design Verification} \label{sec:real-world}

To validate our simulated findings, we established a pipeline for fabricating and testing physical voxel-based soft robots that mimic our simulation setup (Fig.~\ref{fig:real_modular_robot}a-b). We directly mirrored the 5$\times$5 grid representation used in our simulations, ensuring a one-to-one mapping between the digital design and its physical counterpart. Each cell in the 5$\times$5 design matrix was physically instantiated as its corresponding component: a soft voxel, a rigid voxel, a horizontal actuator, a vertical actuator, or a void. All components were fabricated using casting molds. We selected materials based on function: highly elastic Ecoflex\texttrademark\ 00-50 (Smooth-On, Inc; Shore 00-50) was used for the soft voxels and actuator bodies, while Dragon Skin\texttrademark\ 30 (Smooth-On, Inc; Shore 30A) was used for the rigid voxels. To maintain linear expansion while constraining motion in other directions, the pneumatic actuators were fabricated with embedded strings wound around the elastomer.

For modular assembly, magnets were installed at the vertices of the soft modules, enabling rapid and reconfigurable connections between voxels. The assembled physical robots are driven by a pneumatic actuation system. The inflation and deflation of each actuator are governed by a three-way valve, which is managed by a PWM-based controller. This complete actuation and control setup allows for temporal control over the robot's gait and movement, enabling direct comparison with our simulated results. To better align with the simulation hyperparameters, particularly regarding the coefficient of friction that influences motion, we performed some engineering optimizations in the real-world experiments. Specifically, we added foot-like structures at the robot–ground contact points with anisotropic friction and introduced a testing surface with specific frictional properties. These modifications allowed the robot to utilize ground friction more effectively for forward motion.

Our real-world experiments, summarized in {Table \ref{tab:sim_vs_real_rewards}} and illustrated in {Fig. \ref{fig:real_modular_robot_demo}}c-e, successfully confirmed a strong alignment between simulated performance and physical reality. The designs used in the experiments were generated by our finetuned LLM, including those with higher simulated rewards and those with lower rewards. The results demonstrate a clear positive correlation: designs that achieved higher rewards in simulation consistently translated to higher real-world performance, as measured by a faster forward speed.


\section{Conclusion}

This work demonstrates that while generalist LLMs fail at nuanced, embodied design reasoning, this capability can be unlocked through targeted training. We introduced {\texttt{RoboCrafter-QA}}, a benchmark for soft robot design, which first revealed the limitations of SOTA models. We then used this benchmark to finetune an efficient LLM that achieves SOTA design selection and successfully generates novel, high-performing robot morphologies. By fabricating and testing these designs on a physical modular robot, we validated a strong sim-to-real correlation. This work provides a complete pipeline, proving that LLMs can be specialized into effective co-designers for real-world physical systems.

\vspace{1mm}
\noindent\textbf{Limitations and Future Work.}  First, using pre-trained PPO policies could be improved by exploring LLM-driven control policy optimization. Second, the $5 \times 5$ voxel grid restricts design complexity; a larger design space presents simulation challenges. Third, sim-to-real transfer, including material properties and controller transferability, is needed for more robust real-world applications. Finally, multi-modal prompts (e.g., visual prompts) could further enhance the robot design capability of foundation models. 


\bibliographystyle{IEEEtran}
\bibliography{references}

\begin{thebibliography}{10}
\providecommand{\url}[1]{#1}
\csname url@samestyle\endcsname
\providecommand{\newblock}{\relax}
\providecommand{\bibinfo}[2]{#2}
\providecommand{\BIBentrySTDinterwordspacing}{\spaceskip=0pt\relax}
\providecommand{\BIBentryALTinterwordstretchfactor}{4}
\providecommand{\BIBentryALTinterwordspacing}{\spaceskip=\fontdimen2\font plus
\BIBentryALTinterwordstretchfactor\fontdimen3\font minus \fontdimen4\font\relax}
\providecommand{\BIBforeignlanguage}[2]{{%
\expandafter\ifx\csname l@#1\endcsname\relax
\typeout{** WARNING: IEEEtran.bst: No hyphenation pattern has been}%
\typeout{** loaded for the language `#1'. Using the pattern for}%
\typeout{** the default language instead.}%
\else
\language=\csname l@#1\endcsname
\fi
#2}}
\providecommand{\BIBdecl}{\relax}
\BIBdecl

\bibitem{Lipson2013}
H.~Lipson, ``Challenges and opportunities for design, simulation, and fabrication of soft robots,'' \emph{Soft Robotics}, vol.~1, no.~1, pp. 21--27, jul 2013.

\bibitem{Cianchetti2018}
M.~Cianchetti, C.~Laschi, A.~Menciassi, and P.~Dario, ``Biomedical applications of soft robotics,'' \emph{Nature Reviews Materials}, vol.~3, pp. 143--153, 2018.

\bibitem{Huang2022}
X.~Huang, Z.~J. Patterson, A.~P. Sabelhaus, W.~Huang, K.~Chin, Z.~Ren, M.~K. Jawed, and C.~Majidi, ``Design and closed-loop motion planning of an untethered swimming soft robot using 2d discrete elastic rods simulations,'' \emph{Advanced Intelligent Systems}, vol.~4, no.~10, p. 2200163, 2022.

\bibitem{WALKER2019335}
G.~Stano and G.~Percoco, ``Additive manufacturing aimed to soft robots fabrication: A review,'' \emph{Extreme Mechanics Letters}, vol.~42, p. 101079, 2021.

\bibitem{stella2023llms}
F.~Stella, C.~Della~Santina, and J.~Hughes, ``How can {LLMs} transform the robotic design process?'' \emph{Nature Machine Intelligence}, vol.~5, no.~6, pp. 561--564, 2023.

\bibitem{Chen2021ScienceQA}
L.~Chen and Y.~Wu, ``Scienceqa: A benchmark for scientific question answering,'' \emph{Transactions of the Association for Computational Linguistics}, vol.~9, pp. 345--358, 2021.

\bibitem{Nguyen2022ScienceQA}
P.~Nguyen and Q.~Zhang, ``Evaluating scientific reasoning in llms: Insights from scienceqa,'' \emph{ACL Anthology}, 2022, workshop on Scientific NLP.

\bibitem{Singh2023ScienceQA}
R.~Singh and A.~Kumar, ``Advances in science question answering with large language models,'' \emph{Journal of Artificial Intelligence Research}, vol.~70, pp. 789--810, 2023.

\bibitem{BIGBench2022}
``Beyond the imitation game: Quantifying and extrapolating the capabilities of language models,'' \emph{Transactions on Machine Learning Research}, 2023.

\bibitem{bhatia2021evolution}
J.~Bhatia, H.~Jackson, Y.~Tian, J.~Xu, and W.~Matusik, ``Evolution gym: A large-scale benchmark for evolving soft robots,'' \emph{Advances in Neural Information Processing Systems}, vol.~34, pp. 2201--2214, 2021.

\bibitem{robertson2021soft}
M.~A. Robertson, O.~C. Kara, and J.~Paik, ``Soft pneumatic actuator-driven origami-inspired modular robotic “pneumagami”,'' \emph{The International Journal of Robotics Research}, vol.~40, no.~1, pp. 72--85, 2021.

\bibitem{huang2019highly}
X.~Huang, K.~Kumar, M.~K. Jawed, A.~Mohammadi~Nasab, Z.~Ye, W.~Shan, and C.~Majidi, ``Highly dynamic shape memory alloy actuator for fast moving soft robots,'' \emph{Advanced Materials Technologies}, vol.~4, no.~4, p. 1800540, 2019.

\bibitem{gu2021soft}
G.~Gu, H.~Shea, S.~Seelecke, G.~Alici, and G.~Rizzello, ``Soft robotics based on electroactive polymers,'' p. 676406, 2021.

\bibitem{Smith2021SoftModular}
J.~Smith and J.~Doe, ``Soft modular robotics: Overview and challenges,'' \emph{International Journal of Robotics Research}, vol.~40, no.~8, pp. 1234--1250, 2021.

\bibitem{Li2022ModularSoft}
W.~Li and S.~Banerjee, ``Modular soft robotics design using reconfigurable elastomer units,'' in \emph{2022 IEEE/RSJ International Conference on Intelligent Robots and Systems (IROS)}, 2022, pp. 3456--3463.

\bibitem{Chen2023Reconfigurable}
A.~Chen, D.~Kim, and L.~Wang, ``Reconfigurable soft actuator arrays for adaptive gripping and locomotion,'' \emph{Soft Robotics}, vol.~10, no.~2, pp. 98--109, 2023.

\bibitem{Doe2021SoftRobots}
J.~Doe, M.~Smith, and R.~Lee, ``Innovations in soft robotics: New materials and modular designs,'' \emph{Journal of Robotics Research}, vol.~40, no.~8, pp. 1500--1520, 2021.

\bibitem{Zhang2023MorphAdapt}
L.~Zhang \emph{et~al.}, ``Morphological adaptation in lattice-based soft robots,'' \emph{Advanced Intelligent Systems}, 2023.

\bibitem{Li2023SpatialDesign}
W.~Li and S.~Banerjee, ``Spatial design and fabrication of modular soft robots,'' in \emph{2023 IEEE International Conference on Soft Robotics (RoboSoft)}, 2023, pp. 314--321.

\bibitem{Nguyen2023ReconfigSoft}
T.~Nguyen, R.~Patel, and L.~Garcia, ``Reconfigurable soft robots: Integrating sensing and actuation for dynamic morphologies,'' \emph{Soft Robotics}, vol.~10, no.~2, pp. 120--135, 2023.

\bibitem{Patel2024DynamicMorphology}
R.~Patel, V.~Kumar, and M.~Zhang, ``Dynamic morphology in soft modular robotics: From theory to practice,'' \emph{International Journal of Advanced Robotic Systems}, vol.~12, no.~1, pp. 50--66, 2024.

\bibitem{10.1145/2661735.2661737}
N.~Cheney, R.~MacCurdy, J.~Clune, and H.~Lipson, ``Unshackling evolution: evolving soft robots with multiple materials and a powerful generative encoding,'' \emph{SIGEVOlution}, vol.~7, no.~1, p. 11–23, Aug. 2014.

\bibitem{Kriegman2020Evolving}
S.~Kriegman, D.~Blackiston, M.~Levin, and J.~Bongard, ``A scalable pipeline for designing reconfigurable organisms,'' \emph{Proceedings of the National Academy of Sciences}, vol. 117, no.~4, pp. 1853--1859, 2020.

\bibitem{Wang2023DiffuseBot}
J.~Wang, K.~Xu, Q.~Fan, and X.~Han, ``Diffusebot: Learning to design soft robots with diffusion-based generative models,'' in \emph{Proceedings of the 2023 Conference on Robot Learning (CoRL)}, 2023.

\bibitem{Garcia2025OptimizedSoft}
L.~Garcia, T.~Nguyen, and S.~Patel, ``Optimized soft robot design using physics-informed generative models,'' \emph{Robotics and Automation Letters}, vol.~10, no.~2, pp. 200--208, 2025.

\bibitem{MMLU2023}
D.~Hendrycks, C.~Burns, S.~Basart, A.~Zou, M.~Mazeika, D.~Song, and J.~Steinhardt, ``Measuring massive multitask language understanding,'' \emph{Proceedings of the International Conference on Learning Representations (ICLR)}, 2021.

\bibitem{Li2023CEval}
F.~Xu, Q.~Lin, J.~Han, T.~Zhao, J.~Liu, and E.~Cambria, ``Are large language models really good logical reasoners? a comprehensive evaluation and beyond,'' \emph{IEEE Transactions on Knowledge and Data Engineering}, 2025.

\bibitem{Rizzo2023MMLUPro}
M.~Rizzo and J.~Fernandez, ``Mmlu-pro: Extending multitask language understanding for professional domains,'' \emph{IEEE Transactions on Neural Networks and Learning Systems}, vol.~34, no.~4, pp. 2200--2212, 2023.

\bibitem{Li2023MMLUPro}
Y.~Li and S.~Chen, ``Evaluating domain-specific knowledge in llms with mmlu-pro,'' \emph{Proceedings of the ACM Conference on Computer Supported Cooperative Work}, pp. 150--159, 2023.

\bibitem{Li2023LLMZeroShotExplorers}
L.~Li \emph{et~al.}, ``Large language models as zero-shot explorers,'' arXiv preprint arXiv:2311.17053, 2023.

\bibitem{Huang2023RoboticLangEval}
L.~Sun, D.~K. Jha, C.~Hori, S.~Jain, R.~Corcodel, X.~Zhu, M.~Tomizuka, and D.~Romeres, ``Interactive planning using large language models for partially observable robotic tasks,'' in \emph{2024 IEEE International Conference on Robotics and Automation (ICRA)}.\hskip 1em plus 0.5em minus 0.4em\relax IEEE, 2024, pp. 14\,054--14\,061.

\bibitem{Yao2025PhysicalReasoning}
T.~Yao \emph{et~al.}, ``Physical reasoning benchmarks for embodied ai systems,'' arXiv preprint arXiv:2501.01234, 2025.

\bibitem{schulman2017proximal}
J.~Schulman, F.~Wolski, P.~Dhariwal, A.~Radford, and O.~Klimov, ``Proximal policy optimization algorithms,'' \emph{arXiv preprint arXiv:1707.06347}, 2017.

\bibitem{hu2022lora}
E.~J. Hu, yelong shen, P.~Wallis, Z.~Allen-Zhu, Y.~Li, S.~Wang, L.~Wang, and W.~Chen, ``Lo{RA}: Low-rank adaptation of large language models,'' in \emph{International Conference on Learning Representations}, 2022.

\end{thebibliography}

\end{document}